\documentclass[10pt,twocolumn,letterpaper]{article}

\usepackage[pagenumbers]{cvpr} 

\usepackage{graphicx}
\usepackage{amsmath}
\usepackage{amssymb}
\usepackage{booktabs}
\usepackage[accsupp]{axessibility}  

\usepackage[pagebackref,breaklinks,colorlinks]{hyperref}
\usepackage[margin=6pt,font=small,labelfont=bf,labelsep=endash,tableposition=top]{caption}

\usepackage[capitalize]{cleveref}
\crefname{section}{Sec.}{Secs.}
\Crefname{section}{Section}{Sections}
\Crefname{table}{Table}{Tables}
\crefname{table}{Tab.}{Tabs.}

\usepackage{booktabs}
\usepackage{multirow}
\usepackage{color}
\usepackage{eucal}
\usepackage{overpic}
\usepackage{bm}
\usepackage[dvipsnames]{xcolor}
\usepackage{color}

\def\cvprPaperID{9383} 
\def\confName{CVPR}
\def\confYear{2023}

\begin{document}


\title{3DPPE: 3D Point Positional Encoding for \\ Multi-Camera 3D Object Detection Transformers}

\author{Changyong Shu$^{1}$\footnotemark[1] \quad Jiajun Deng$^{2}$\footnotemark[1] \quad Fisher Yu$^{3}$ \quad Yifan Liu$^{4}$\footnotemark[2] \\
$^1$Houmo AI, $^2$University of Sydney, $^3$ETH Zürich, $^4$University of Adelaide, \\
     {\tt\small changyong.shu89@gmail.com}, {\tt\small jiajun.deng@sydney.edu.au}\\
      {\tt\small fisheryu@ethz.ch, yifan.liu04@adelaide.edu.au
 }
}

\maketitle
\renewcommand{\thefootnote}{\fnsymbol{footnote}} 
\footnotetext[1]{These authors contributed equally to this work.} 
\footnotetext[2]{Corresponding authors.} 

\begin{abstract}

Transformer-based methods have swept the benchmarks on 2D and 3D detection on images.  Because tokenization before the attention mechanism drops the spatial information, positional encoding becomes critical for those methods.  Recent works found that encodings based on samples of the 3D viewing rays can significantly improve the quality of multi-camera 3D object detection.  We hypothesize that 3D point locations can provide more information than rays.  Therefore, we introduce 3D point positional encoding, 3DPPE, to the 3D detection Transformer decoder. Although 3D measurements are not available at the inference time of monocular 3D object detection, 3DPPE uses predicted depth to approximate the real point positions.  Our hybrid-depth module combines direct and categorical depth to estimate the refined depth of each pixel. Despite the approximation, 3DPPE achieves 46.0 mAP and 51.4 NDS on the competitive nuScenes dataset, significantly outperforming encodings based on ray samples. We will make the codes available for further investigation.

\end{abstract}

\section{Introduction}
3D object detection is a vital component of autonomous driving perception systems. 
Particularly, image-based 3D object detection has received increasing attention from both academia and industry due to its lower cost compared to LiDAR-dependent solutions.
Despite the fact that autonomous driving vehicles are equipped with multiple cameras, early attempts at image-based 3D object detection, as seen in previous works~\cite{park2021pseudo,fcos3d}, focus on monocular detection and combine the detection results from multiple cameras.
This kind of solution is unable to make use of correspondence in the overlapping area of adjacent cameras, and the paradigm to individually detect objects in each view involves a large computational overhead. 
Alternatively, a group of recent studies~\cite{bevdet,bevdet4d,bevdepth,sts} follow the paradigm of Lift-Splat-Shoot (LSS)~\cite{lss} to first transform multi-camera images to unified bird-eye-view (BEV) representation in parallel and then perform object detection on the BEV representation.
However, such ill-posed view transformation inevitably causes error accumulation, which further affects the accuracy of 3D object detection.

\begin{figure}
\centering
		\includegraphics[width=1.0\linewidth]{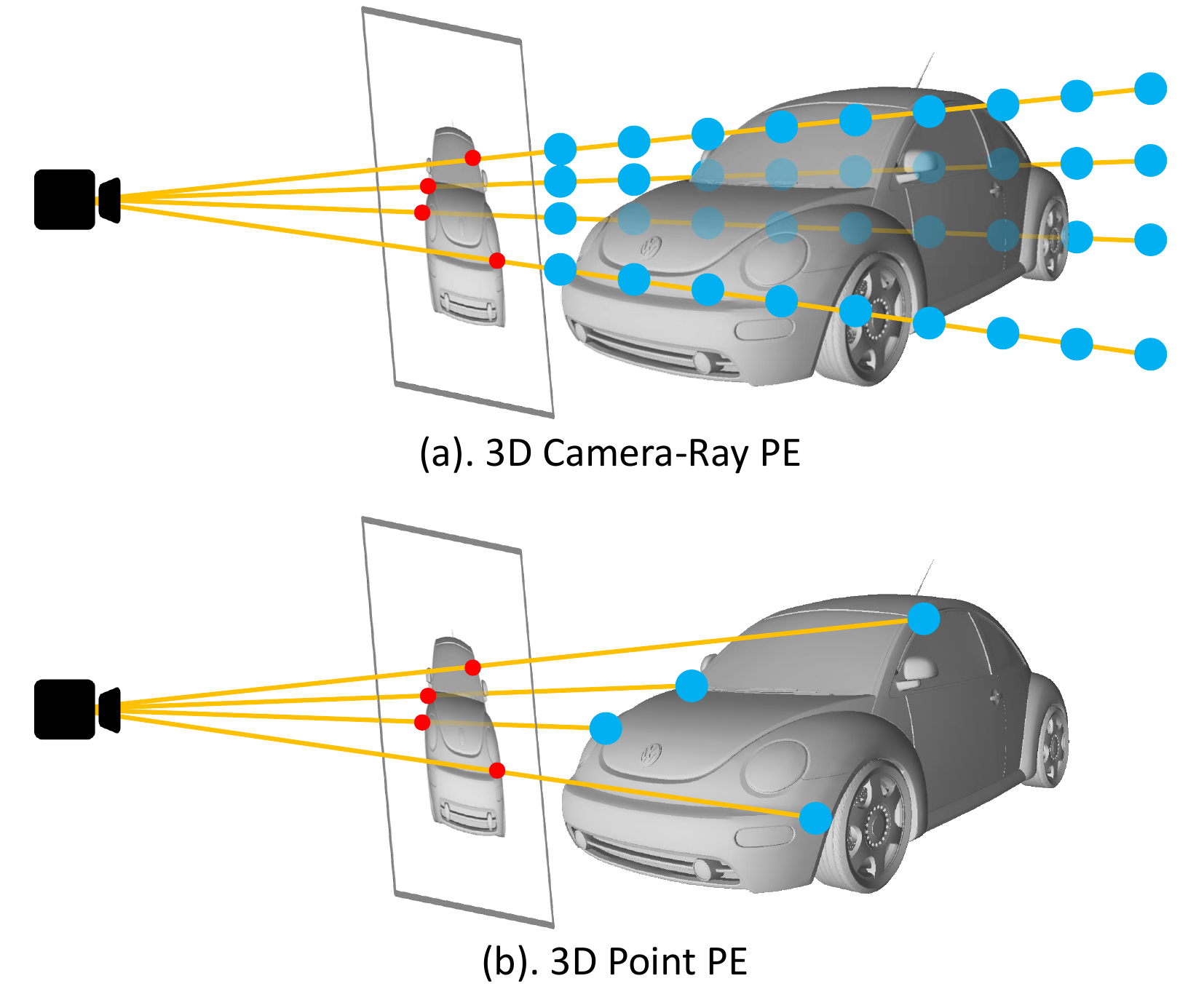}
	
	\caption{
	An illustration of (a) 3D camera-ray positional encoding (PE) and (b) our proposed 3D point PE. The 3D camera-ray PE represents camera-ray information by determining the positions of a set number of discrete points along the direction from the camera optical center to the image plane pixel. This encoding approach is coarse-grained. On the other hand, the 3D point PE provides more precise position information by encoding the location of a single point with an estimated depth. In the figure, four pixels are randomly selected to demonstrate the methods.
    }
	\label{fig:fig1}
\end{figure}

At the same time, the transformer-based (DETR-like)~\cite{detr} scheme has also been explored in this field. Typically, the methods following this scheme \cite{futr3d,petrv2,petr,detr3d,bevformer} utilizes a set of learnable 3D object query to iteratively interact with multi-view 2D features, and further perform 3D object detection without explicit view transformation.
Within the transformer-based methods, there are two general ways to enable the interaction of 3D queries and 2D image features, \emph{i.e.}, projection-based and position-encoding-based.
The former one projects 3D queries into the 2D image plane~\cite{detr3d,bevformer} for feature sampling, which requires extra deployment efforts. 
Moreover, such a sampling procedure only extracts local features, failing to make use of global coherence for improving 3D object detection. 
The other way, as first introduced in PETR~\cite{petr}, integrates the 3D information into 2D image features by positional encoding. 
With 3D positional encoding (PE), 2D image features can be directly exploited by 3D queries, without extra projection efforts. 



Enhancement of the 3D PE is anticipated to result in more precise 3D object detection.
Despite effectiveness, the mechanism and design options of 3D PE in previous methods have not been fully explored. The typical 3D PE is the 3D camera-ray PE, as shown in Figure.~\ref{fig:fig1} (a).
It encodes the ray direction starting from the camera's optical center to the pixel on the image plane. However, the ray direction only provides coarse localization information for the 2D image feature without the depth prior. Moreover, as the object query is embedded from the randomly initialized 3D reference point, the inconsistent embedding space for the reference point and camera-ray PE further hampers the effectiveness of the attention mechanism in the transformer decoder. Thus, reformulating a new 3D positional encoding with depth prior to localize the 2D feature and unify representation for both image feature and object query is still a legacy issue.

In this work, we explore an alternative 3D PE paradigm to ameliorate the aforementioned problem. Formally, we introduce 3D point positional encoding (3DPPE) to improve transformer-based multi-camera 3D object detection. As illustrated in Figure.~\ref{fig:fig1} (b), 3DPPE improves the camera-ray 3D PE by involving depth prior. Moreover, we find that 3D point PE not merely avoids the defects above, but also can provide better representative similarity (shown in Figure.~\ref{fig:fig5}). Specifically, in 3DPPE, we first devise a hybrid-depth module that combines direct and categorical ones to estimate the refined depth of each pixel.
Then, we transform the pixels to 3D points via the camera parameters and predicted depth.
The resulting 3D points are sequentially sent to a position encoder for 3D point PE. 
Particularly, we exploit a shared position encoder for the transformed 3D points and reference points to develop a unified embedding space. 

We conduct extensive experiments to demonstrate the advantages of our proposed 3DPPE on challenging NuScene benchmarks. With the proposed 3D point positional encoding, our proposed 3DPPE can improve the camera-ray-based encoding by 1.9\% mAP and 1.0\% NDS. 


\section{Related Work}

\noindent \textbf{Transformer-based object detection.}
Object detection has been an active research topic in computer vision for several decades. Traditional object detection approaches, such as sliding window-based methods (one-stage) and region-based (two-stage) methods, have achieved significant progress in recent years. However, these methods generally rely on hand-designed components, such as non-maximum-suppression (NMS) or anchor generation.
DETR \cite{detr} is a pioneering work that introduces the transformer-based framework to solve object detection as a set prediction problem, eliminating the need for heuristic target assignment and extra post-processing like non-maximum suppression (NMS). Deformable DETR \cite{deformabledetr} improves DETR by introducing deformable attention and multi-level image features to ameliorate the slow convergence problem and to improve the poor detection performance for small objects.
Two-stage schemes \cite{deformabledetr,efficientdetr,tsp} use the top-k scoring region proposals to initialize the object queries for convergence acceleration.
\cite{anchordetr,dabdetr,acceleratingdetr} use anchor points or anchor boxes to generate object queries, which provide explicit positional priors.
SMCA \cite{smca} and Conditional DETR \cite{conditionaldetr} enhance the cross-attention mechanisms by leveraging the spatial information in the decoder embedding. DN-DETR \cite{dndetr} and its variant DINO \cite{dino} incorporate denoising techniques to ameliorate the instability problem of bipartite graph matching.
\begin{figure*}
\centering
		\includegraphics[width=1.0\linewidth]{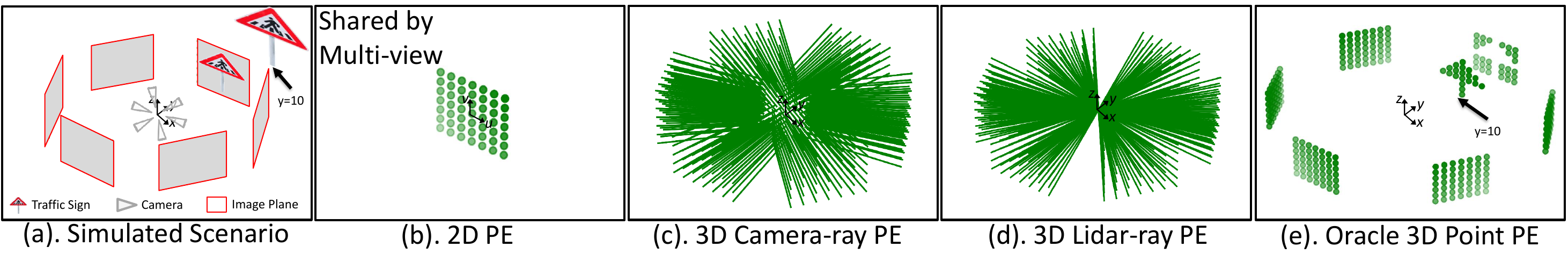}
	\caption{Illustration of different positional encoding in the surround-view system. (a) presents the simulated scenario with 6 cameras, only one traffic sign appears in front view (the distance is set as 10m), rather than other cameras. Comparing all the PE from (b) to (e), only the oracle 3D point PE can encode the precise 3D point location of the object. Best viewed in color.
	}
	\label{fig:different_position_embedding}
\end{figure*}

\noindent \textbf{Multi-camera 3D object detection.} Previous works on multi-camera object detection have typically used monocular detection to process each view separately, followed by post-processing to merge the results into a unified coordinate system. However, this approach is limited in its ability to utilize information from multiple views simultaneously and can lead to missed detections, particularly for truncated objects. A promising paradigm has emerged recently that converts multi-camera features from perspective view to bird's-eye view (BEV) and performs object detection under BEV. Two representative approaches within this paradigm are LSS-based and transformer-based.

The LSS-based methods, such BEVDet \cite{bevdet} and BEVDet4D \cite{bevdet4d}, are effective approaches for converting multi-camera features into a dense bird's-eye view (BEV) representation using LSS \cite{lss}. Specifically, these methods predict the categorical depth distribution of each pixel in the image feature map to generate the dense BEV representation, which can provide comprehensive information for 3D object detection.  Following methods, such as BEVDepth \cite{bevdepth} and STS \cite{sts} explicitly introduce a sub-network for depth estimation to refine the depth prediction.


Among the transformer-based methods, BEVFormer \cite{bevformer} constructs a dense BEV representation using a set of grid-shaped BEV queries to aggregate spatial and temporal features. DETR3D \cite{detr3d} samples 2D image features by projecting 3D reference points generated by object queries onto all views. The PETR series \cite{petr, petrv2} proposes the 3D position-encoding (PE) to transform the image features into 3D position-aware features, which can be directly interacted with object queries in 3D space. Following PETR series, Focal-PETR \cite{wang2022focal} utilizes instance-guided supervision and spatial alignment module to adaptively focus object queries on discriminative foreground regions;
MV2D \cite{wang2023object} generates a dynamic object query from 2D detector result, and one object query aggregates the feature from its corresponding 2D bounding box region.

In this paper, we follow PETR to perform 3D object detection with the transformer-based paradigm. However, in contrast to the previous approaches \cite{wang2022focal, wang2023object} that leverage 2D prior to improving 3D object detection, we devote our main efforts on investigate the 3D positional encoding, which has been rarely studied in the literature.

\section{Preliminary of Positional Encoding}
\subsection{Ray-based Positional Encoding}

The PETR series methods, \emph{i.e.}, PETR~\cite{petr} and PETRv2~\cite{petrv2}, introduce a technique for multi-camera 3D object detection by encoding 3D coordinate information into multi-camera image features. This approach allows for the production of 3D position-aware features, which can improve the accuracy of object detection in 3D space. Specifically, PETR and PETRv2 obtain the 3D coordinate information from a set of points along the camera ray, namely camera-ray PE. 
Given the depth range $R_D=[D_\text{min}, D_\text{max}]$, camera-ray PE first divides the depth into $N_D$ bins via linear-increasing discretization (LID)~\cite{petr}. The center of each bin is exploited to represent the corresponding bin, and thus the 3D position information of a pixel is represented as $N_D$ points along the camera-ray direction. After that, by utilizing the extrinsic and intrinsic parameters of the camera, points corresponding to different camera views are transformed into a unified coordinate system. For each pixel, the camera-ray points are concatenated together and fed into an embedding layer for positional encoding.
We perform further analysis on the ray-based positional encoding in PETR and PETRv2 in the supplementary material.

\subsection{3D Point Positional Encoding} %
\label{sec:3.3}

For optimal accuracy in positional encoding, it is important to have access to the true 3D position of a point on a 2D plane, as demonstrated in Figure.~\ref{fig:different_position_embedding}-(e). In contrast, camera-ray PE encodes the direction from the camera's optical center to the pixel on the image plane, while LiDAR-ray PE depicts the orientation from the LiDAR center to the 3D point. While both ray PEs encode direction, they cannot accurately determine the 3D location without precise depth information.

\begin{figure*}
\centering
	\includegraphics[width=1.0\linewidth]{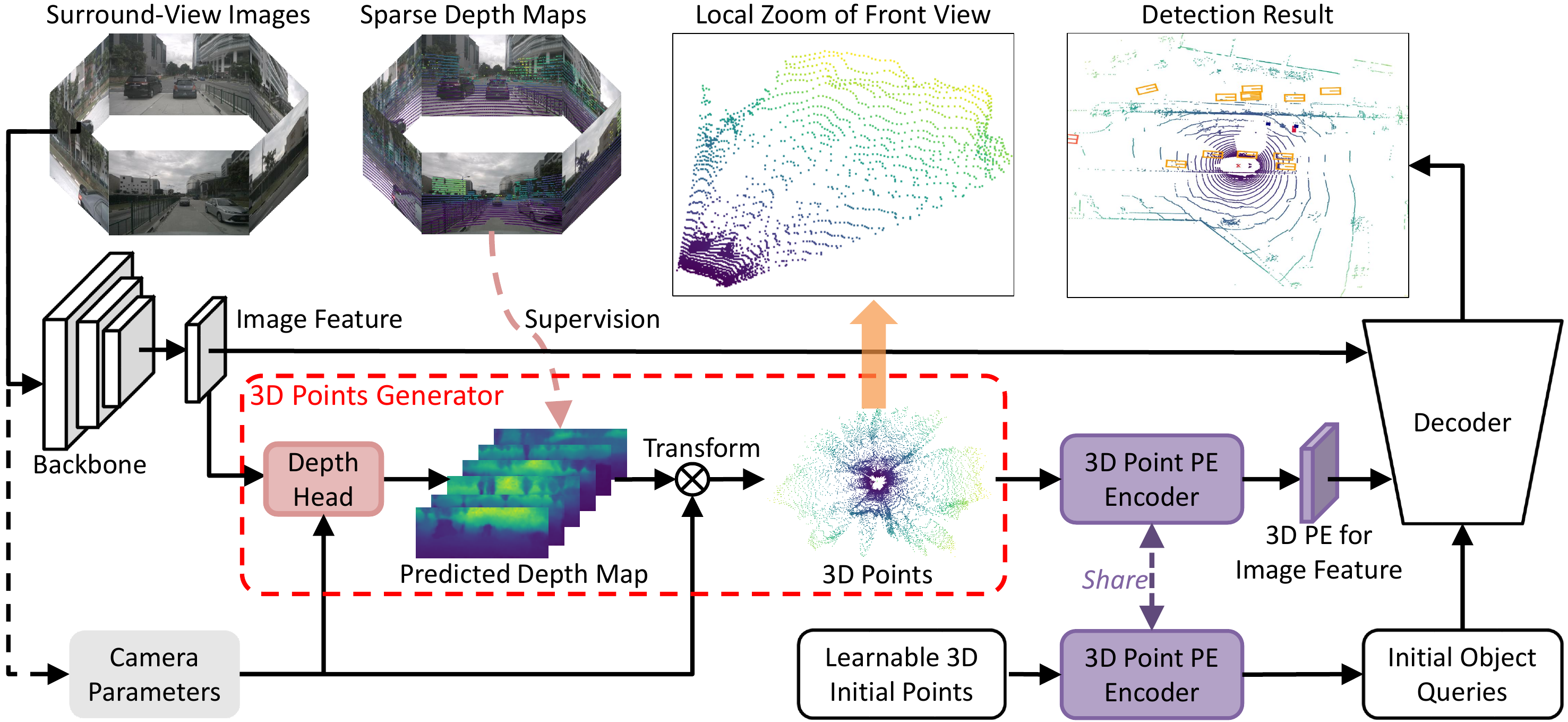}
	\caption{The overall architecture of our proposed pipeline for method. Best viewed in color and zoom. 
	The plot on the up shows intuitive illustration of surround-view images, sparse point projection, local zoom of front view and detection result.
	In local zoom of front view,
	the two dense point blocks located in left bottom corner indicate the two car object in the front view.
	The plot on the bottom shows the detail component of our method.
	}
	\label{fig:overview}
\end{figure*}

To confirm that accurate 3D point positioning can enhance detection performance, we project 2D features into 3D space using the ground truth depth of a 2D image. Specifically, we project 3D point clouds onto surround-view images to generate a sparse depth map, and then use depth completion \cite{ku2018defense} to obtain the ground truth (GT) dense depth. We compare the performance of different positional encoding settings in Figure.~\ref{fig:different_position_embedding} and list the results in Table.~\ref{tab:different_pe}.  All experiments are compared following the same training scheme.
Compared to 3D camera-ray PE,
2D PE scores worst due to its complete inability to multi-camera distinction.
3D LiDAR-ray PE can achieve on-par or inferior performance depending on the fixed depth $d$.
As for the 3D point PE with ground truth depth named Oracle 3D Point, a significant improvement is achieved with 6.7\% NDS, 10.9\% mAP, and 18.7\% mATE respectively,
which verify that 
{\it \textbf{the 3D PE encoded from precise 3D point location is the key to improve multi-camera 3D object detection}}.

\begin{table}[!htb] 
    \caption{
    Performance comparison of different PE settings. 
    Comparing all PEs listed in row 1, only the oracle 3D pint PE can encode precise 3D point location of objects.
    }
    \vspace{0.25cm}
    \setlength{\tabcolsep}{5.5mm}
    \centering
    \footnotesize
    \begin{tabular}{l|p{0.6cm}p{0.6cm}c}
    \toprule[1.5pt]
     PE &  NDS$\uparrow$ & mAP$\uparrow$ & mATE$\downarrow$ \\
    \noalign{\smallskip}
    \hline
    \noalign{\smallskip}
    Camera-ray&0.337&0.274&0.852  \\
    \hline
    2D&0.193&0.055&1.209 \\ 
    LiDAR-ray&0.338&0.275&0.849 \\ 
    Oracle 3D Point&\bf{0.404}&\bf{0.383}&\bf{0.665}\\
    \bottomrule[1.5pt]
    \end{tabular}

    \label{tab:different_pe}
\end{table}

As a camera-only system cannot gather ground truth depth information, we introduce a lightweight depth estimation module to substitute for the inaccessible GT depth. A more precise depth estimation results in improved 3D object detection performance. This study illustrates the potential of encoding 2D image features in 3D space with the help of estimated depth information.


\section{Method}
In this section, 
we present how to utilize the proposed unified depth-guided 3D point PE to transform the 2D features
from multi-view images into the 3D space to perform multi-camera 3D object detection. 
We start by giving the architecture overview (Section. \ref{sec:4.1}), 
then depict 3D point generator (Section. \ref{sec:4.2}) 
and 3D point encoder (Section. \ref{sec:4.3}), 
ultimately elaborate 
3D point-aware feature (Section. \ref{sec:4.4})
and 
decoder modification (Section. \ref{sec:4.5}) 
respectively.   

\subsection{Framework Overview}
\label{sec:4.1}
In this work, we present 3D point positional encoding (3DPPE) for transformer-based multi-camera 3D object detection. 
As shown in Figure.~\ref{fig:overview},
We first send $N$ surround-view images $\boldsymbol{I}=\{ I_i \in \mathbb{R}^{3 \times H_{I_{i}} \times W_{I_{i}}}, i=1,2,\dots, N \}$ to backone (e.g. ResNet \cite{resnet}, Swintransformer \cite{swin}) for image features $\boldsymbol{F}=\{ F_i \in \mathbb{R}^{C \times H_{F_{i}} \times W_{F_{i}}}, i=1,2,\dots, N \}$, where $H_{I_i}$ and $W_{I_i}$ is the $i$-th image shape, $H_{F_i}$ and $W_{F_i}$ is the $i$-th feature shape, $C$ is the channel number of $i$-th feature.
Then the image feature $\boldsymbol{F}$ undergo the depth head in 3D point generator for dense depth maps $\boldsymbol{D}=\{ D_i \in \mathbb{R}^{1 \times H_{F_{i}} \times W_{F_{i}}}, i=1,2,\dots, N\}$,
and $\boldsymbol{D}$ is further transferred to the 3D points $P^{\mathtt{3D}}=\{P^{\mathtt{3D}}_i \in \mathbb{R}^{3 \times H_{F_{i}} \times W_{F_{i}}}, i=1,2,\dots,N\}$ via camera parameter.
The shared 3D point PE generator imports the 3D points $\boldsymbol{P}^{3D}$ above to produce the 3D point PE following $\mathtt{PE}=\{ \mathtt{PE}_i \in \mathbb{R}^{C \times H_{F_{i}} \times W_{F_{i}}}, i=1,2,\dots, N\}$ for 2D image feature. The 3D point PE generator also takes in the learnable 3D anchor points for 3D object queries $Q=\{ Q_i \in \mathbb{R}^{C \times 1}, i=1,2,\dots, K\}$, where $PE$ and $Q$ are unified 3D presentation with fine-grained point-aware position in 3D space. 
Finally, the 3D queries can directly interact with the image features supplemented by the 3D point PE in decoder to perform 3D object detection.

\begin{figure}[t]
\centering
      \includegraphics[width=1.05\linewidth]{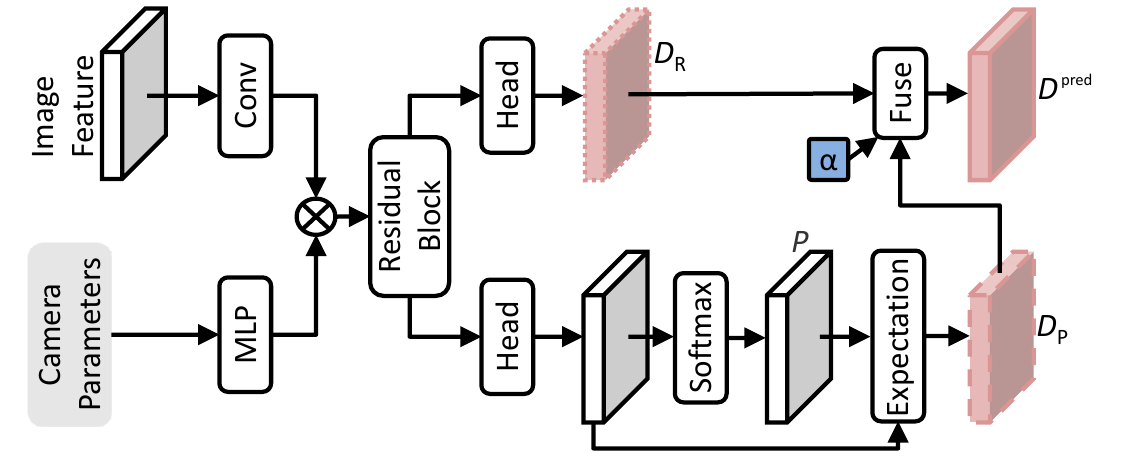}
   \caption{Framework for depth head. The depth estimation $D^{pred}$ is the fusion result of regressed depth $D^R$ and probabilistic depth $D^P$, $\alpha$ is the fusion weight, $P$ is the probabilistic over depth bins.}
   \label{fig:fig4}
\end{figure}

\begin{figure*}
\centering
		\includegraphics[width=1.0\linewidth]{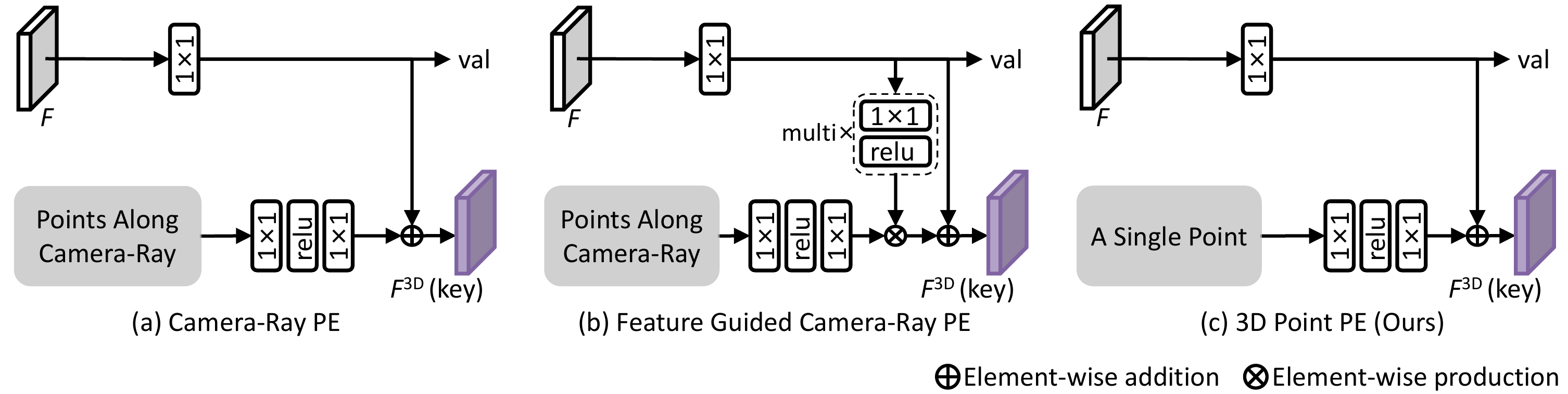}
	\caption{Comparison between (a) camera-ray PE, (b) feature guided camera-ray PE, and (c) our proposed 3D point PE.}
	\label{fig:fig3}
\end{figure*}

\subsection{3D Point Generator}
\label{sec:4.2}
We introduce a depth estimation module to provide dense depth map, and then transfer it to 3D point via camera back-projection.

\noindent
\textbf{Hybrid-Depth Module.} Inspired by BEVdepth \cite{bevdepth} and PGD \cite{pgd}, we design a hybrid-depth module that fuses the directly regressed depth $D^{\mathtt{R}} \in \mathbb{R}^{H_F \times W_F}$ and categorical depth $D^{\mathtt{P}} \in \mathbb{R}^{H_F \times W_F}$ with a learnable weight $\alpha$. The architecture of our proposed hybrid-depth module is illustrated in Figure.~\ref{fig:fig4}.  For a given depth range $[d_{\mathtt{min}},d_{\mathtt{max}}]$ with identical interval $d_{\Delta}$, we could get $N_\mathtt{D}=\frac{d_{\mathtt{max}}-d_{\mathtt{m
in}}}{d_{\Delta}}$ discrete depth bins: $\textbf{D}=\{d_1, d_2, \dots, d_{N_\mathtt{D}}\}$. Instead of directly regressing the relative depth, the probabilistic $P$ over these depth bins could be generated for each pixel, $P \in \mathbb{R}^{N_\mathtt{D}\times H_F \times W_F}$. Thus, the pixel depth can be formulated as:
\begin{align}
 D^\mathtt{P} = \sum_{i=1}^{N_\mathtt{D}}P_{u,v,i} \times d_{i}.
\end{align}

The ultimate depth estimation $D^{\mathtt{pred}}$ comes from the fusion result of $D^{\mathtt{R}}$ and $D^{\mathtt{P}}$ above:
\begin{align}
  D^{\mathtt{pred}} = {\alpha}D^{\mathtt{R}} + (1 - \alpha)D^{\mathtt{P}},
  \label{eq:integrated depth}
\end{align}
where $\alpha$ is the learnable fusion weight for proportion balance. To achieve reliable depth prediction, predicted depth are supervised by projected depth $D^{\mathtt{gt}}$ from point cloud, and smooth L1 loss \cite{fcos3d,pgd} and distribution focal loss \cite{gfl} are utilized:
\begin{align}
    \begin{split}
    L_{\mathtt{depth}} = &\lambda_{\mathtt{sm}} L_{\mathtt{smooth}-\mathtt{L1}}(D^{\mathtt{pred}}, D^{\mathtt{gt}}) \\
                & + \lambda_{\mathtt{dfl}} L_{\mathtt{dfl}}(D^{\mathtt{pred}}, D^{\mathtt{gt}}, \textbf{D}),
    \end{split}
\label{eq:integrated depth}
\end{align}
where $\lambda_{\mathtt{sm}}$ and $\lambda_{\mathtt{dfl}}$ is the hyper-parameters. The $L_{\mathtt{dfl}}$ 
aims to enlarge the probabilities of nearest two bins 
${d^{i}}$ and ${d^{i+1}}$
around the ground truth ${D^{\mathtt{gt}}}$ 
($d_{i} < {D^{gt}} < d_{i+1}$) 
for efficient learning:
\begin{align}
    \begin{split}
	L_{\mathtt{dfl}}(D^{\mathtt{pred}}, D^{\mathtt{gt}}, \textbf{D}) = &-\frac{d_{i+1} -D^{\mathtt{gt}}}{d_{\Delta}}log(P_{i}) \\
	                                        &-\frac{D^{\mathtt{gt}} - d_{i}}{d_{\Delta}}log(P_{i+1}).
	\end{split}
    \label{eq:hinge_loss}
\end{align}

\noindent
\textbf{Coordinate Transformation from 2D to 3D.}
We transfer the 2D pixels in surround view to 3D point 
$P^{\mathtt{3D}}$ 
in the LiDAR coordinate system via the camera parameters. This process can be formulated as follow:
\begin{equation}
  \begin{bmatrix}
    P^{\mathtt{3D}}_{i}[0,u,v] \\
    P^{\mathtt{3D}}_{i}[1,u,v] \\
    P^{\mathtt{3D}}_{i}[2,u,v] \\
    \end{bmatrix} = R_iK_i^{-1}D^{\mathtt{pred}}_{i}[u,v]\begin{bmatrix}
    u \\
    v \\
    1 \\
    \end{bmatrix}+T_i.
\label{eq:2.5D-to-3D}
\end{equation}
where $D^{\mathtt{pred}}_{i}[u,v]$\footnote{The $[\dots]$ in the whole paper denotes the tensor slice in pytorch.} is the predicted depth of 2D pixel $(u,v)$, 
$P^{\mathtt{3D}}_{i}[0,u,v]$, $P^{\mathtt{3D}}_{i}[1,u,v]$ and $P^{\mathtt{3D}}_{i}[2,u,v]$ are the x-axis, y-axis and z-axis coordinate of the correspond 3D point for 2D pixel $(u,v)$ in $i$-th camera. $K_i \in \mathbb{R}^{3 \times 3}$ is $i$-th camera intrinsic matrix, $R_i \in \mathbb{R}^{3 \times 3}$ and $T_i \in \mathbb{R}^{3 \times 1}$ are the rotation and translation matrix from the camera coordinate system of $i$-th view to LiDAR coordinate system.

Setting the region of 3D perception space $[x_{\mathtt{max}}, x_{\mathtt{min}}, $
$y_{\mathtt{max}}, y_{\mathtt{min}}, z_{\mathtt{max}}, z_{\mathtt{min}}]$, the normalization is further conducted on each 3D point:
\begin{equation}\label{eq2}
\left\{
            \begin{array}{lr}
            P^{\mathtt{3D}}_{i}[0,u,v] = (P^{\mathtt{3D}}_{i}[0,u,v]-x_{\mathtt{min}}) / (x_{\mathtt{max}}-x_{\mathtt{min}})\\
            P^{\mathtt{3D}}_{i}[1,u,v] = (P^{\mathtt{3D}}_{i}[1,u,v]-y_{\mathtt{min}}) / (y_{\mathtt{max}}-y_{\mathtt{min}})\\
            P^{\mathtt{3D}}_{i}[2,u,v] = (P^{\mathtt{3D}}_{i}[2,u,v]-z_{\mathtt{min}}) / (z_{\mathtt{max}}-z_{\mathtt{min}}).
             \end{array}
\right.
\end{equation}

\subsection{3D point Encoder}
\label{sec:4.3}
The 3D point 
$P^{\mathtt{3D}}$
is embedded in 3D point PE 
$PE$
via the 3D point encoder:
\begin{equation}
    \begin{split}
    \mathtt{PE}_{i}[:,u,v] = \mathtt{MLP}(\mathtt{Cat}(&\mathtt{Sine}(P^{\mathtt{3D}}_{i}[0,u,v]), \\
                                 &\mathtt{Sine}(P^{\mathtt{3D}}_{i}[1,u,v]), \\
                        &\mathtt{Sine}(P^{\mathtt{3D}}_{i}[2,u,v]))),
    \end{split}
\end{equation}
where the sine/cosine positional encoding function $\mathtt{Sine}$ \cite{detr} maps a $1$-dimensional coordinate value to a $\frac{C}{2}$-dimensional vector, the sequential $\mathtt{Cat}$ operator concatenate $\mathtt{Sine}(P^{\mathtt{3D}}_{i}[0,u,v])$, $\mathtt{Sine}(P^{\mathtt{3D}}_{i}[1,u,v])$ and $\mathtt{Sine}(P^{\mathtt{3D}}_{i}[2,u,v])$ to generate a $\frac{3C}{2}$-dimensional vector, then the $\mathtt{MLP}$ consisted of two linear layers and a RELU activation reduces the vector dimension from $\frac{3C}{2}$ to $C$.

\begin{table*}[t]
\setlength{\tabcolsep}{3.1mm}
\centering
\footnotesize
\caption{\textbf{Performance comparison of 3D object detection on nuScenes val set.} 
$\dagger$ indicates using the pre-trained FCOS3D backbone for model initialization. 
“S” denotes model with a single time stamp input. $\ast$ is trained with CBGS.
}

\label{tab:val_tab}
\begin{tabular}{l|c c|c c| c c c c c}
\toprule[1.5pt]
Methods & Backbone & Resolution & mAP$\uparrow$ & NDS$\uparrow$ & mATE$\downarrow$ & mASE$\downarrow$ & mAOE$\downarrow$ & mAVE$\downarrow$ & mAAE$\downarrow$ \\
\midrule
BEVDet~\cite{bevdet} & Res-50 & $ 704 \times 256 $ & 0.298 & 0.379 & 0.725 & 0.279 & 0.559 & 0.860 & 0.245 \\
BEVDepth-S~\cite{bevdepth} & Res-50 & $ 704 \times 256 $ & 0.315 & 0.367 & 0.702 & 0.271 & 0.621 & 1.042 & 0.315 \\
PETR$\ast$ \cite{petr} & Res-50 & $ 1408 \times 512 $ & 0.339 & 0.403 & 0.748 & 0.273 & 0.539 & 0.907 & 0.203 \\
\midrule
3DPPE$\ast$ & Res-50 & $ 1408 \times 512 $ & \textbf{0.370} & \textbf{0.433} & 0.689 & 0.279 & 0.524 & 0.828 & 0.202 \\

\midrule
\midrule
FCOS3D\cite{fcos3d} & Res-101 & $ 1600 \times 900 $ & 0.295 & 0.372 & 0.806 & 0.268 & 0.511 & 1.315 & 0.170 \\
PGD\cite{pgd} & Res-101 &  $ 1600 \times 900 $ & 0.335 & 0.409 & 0.732 & 0.263 & 0.423 & 1.285 & 0.172 \\
DETR3D$\ast$$\dagger$\cite{detr3d} & Res-101 & $ 1600 \times 900$  & 0.349 & 0.434 & 0.716 & 0.268 & 0.379 & 0.842 & 0.200 \\
BEVFormer-S$\ast$$\dagger$\cite{bevformer} & Res-101 & $ 1600 \times 900 $ & 0.375 & 0.448 & 0.725 & 0.272 & 0.391 & 0.802 & 0.200 \\
Ego3RT$\ast$$\dagger$\cite{lu2022learning} & Res-101 & $ 1600 \times 900 $ & 0.375 & 0.450 & 0.657 & 0.268 & 0.391 & 0.850 & 0.206 \\
SpatialDETR$\ast$$\dagger$\cite{doll2022spatialdetr} & Res-101 & $ 1600 \times 900 $ & 0.351 & 0.425 & 0.772 & 0.274 & 0.395 & 0.847 & 0.217 \\
PETR$\ast$$\dagger$\cite{petr} & Res-101 & $ 1408 \times 512 $ & 0.366 & 0.441 & 0.717 & 0.267 & 0.412 & 0.834 & 0.190 \\
\midrule
3DPPE$\ast$$\dagger$ & Res-101 & $ 1408 \times 512 $ & \textbf{0.391} & \textbf{0.458} & 0.674 & 0.282 & 0.395 & 0.830 & 0.191 \\
\bottomrule[1.5pt]
\end{tabular}
\end{table*}

\begin{table*}[t]
  \setlength{\tabcolsep}{4.3mm}
  \centering
  \footnotesize
  \caption{\textbf{Performance comparison of 3D object detection performance on nuScenes test set.} 
“S” denotes model with a single time stamp input.}
  \vspace{0.25cm}
  \label{test_tab}
  \begin{tabular}{l|c|c c |c c c c c}
  \toprule[1.5pt]
    Methods & Backbone & mAP$\uparrow$ & NDS$\uparrow$ & mATE$\downarrow$ & mASE$\downarrow$ & mAOE$\downarrow$ & mAVE$\downarrow$ & mAAE$\downarrow$ \\
  \midrule
    DD3D\cite{park2021pseudo} & VoV-99 & 0.418 & 0.477 & 0.572 & 0.249 & 0.368 & 1.014 & 0.124 \\
    DETR3D\cite{detr3d} & VoV-99 & 0.412 & 0.479 & 0.641 & 0.255 & 0.394 & 0.845 & 0.133 \\
    Ego3RT\cite{lu2022learning} & VoV-99 & 0.425 & 0.473 & 0.549 & 0.264 & 0.433 & 1.014 & 0.145 \\
    BEVDet\cite{bevdet} & VoV-99 & 0.424 & 0.488 & 0.524 & 0.242 & 0.373 & 0.950 & 0.148 \\
    BEVFormer-S\cite{bevformer} & VoV-99 & 0.435 & 0.495 & 0.589 & 0.254 & 0.402 & 0.842 & 0.131 \\
    SpatialDETR\cite{doll2022spatialdetr} &  VoV-99 & 0.424 & 0.486 & 0.613 & 0.253 & 0.402 & 0.857 & 0.131 \\
    PETR\cite{petr} & VoV-99 & 0.441 & 0.504 & 0.593 & 0.249 & 0.383 & 0.808 & 0.132\\
  \midrule
    3DPPE & VoV-99 & \textbf{0.460} & \textbf{0.514} & 0.569 & 0.255 & 0.394 & 0.796 & 0.138 \\
  
  \bottomrule[1.5pt]
  \end{tabular}
\end{table*}

\subsection{3D Point-Aware Features}
\label{sec:4.4}
Given the resulting 3D point PE for image features above, we add it element-wisely with the image feature $F$ for generating the 3D point-aware features $F^{\mathtt{3D}}$. 
For a better understanding of our proposed 3D point-aware feature, Figure. \ref{fig:fig3} illustrates the difference among ours 3D point-aware feature and 3D position-aware feature in PETR series: (1) the channel dimensionality of point cloud in petr series is $N_{\mathtt{D}} \times 3$, where $N_{\mathtt{D}}$ denotes the depth bin number along camera-ray, and the positional encoding is generated in a ray-aware paradigm. Whereas our method performs the positional encoding in point-aware manner with the channel dimensionality of point cloud reduce to $1 \times 3$ (e.g., the definite depth leading to better locating capability in 3D space). 
(2) Our scheme is compact for explicit motivation, as the function-ambiguity multi-layer modulation for feature-guided 3D PE in PETRv2 is not used.

\subsection{Modification in Decoder}
\label{sec:4.5}
As depicted in Figure. \ref{fig:overview}, the learnable 3D anchor points go through the shared embedding generator used for 2D image features above to produce the 3D point PE $E^Q$ for random initialized object queries $Q$, thus the $E^F$ and $E^Q$ are essentially encoded in the sympatric representation, which further enhances the object queries with precise positioning for indexing the useful 3D point-aware feature and performing the accurate 3D object.


\section{Experiment}
In this section, we first present the main results of our 3DPPE and compare with other state-of-the-art methods on nuScenes dataset in Section.~\ref{sec:exp_main}. Then, in Section.~\ref{sec:exp_ablation}, we conduct extensive ablative experiments to investigate the effectiveness of each component in our proposed method. After that, in Section.~\ref{sec:exp_showcase}, we show the qualitative comparison between our method and the previous ray-based positional encoding. Finally, we discuss further potential improvements of 3DPPE in Section.~\ref{sec:exp_discuss}.
We detail the benchmark and metrics, as well as experimental details, in the supplementary material.

\subsection{Comparison with State-of-the-art Methods} \label{sec:exp_main}
We compare the proposed method with other state-of-the-art multi-camera 3D object detectors on the validation and test sets of nuScenes dataset. All of the reported methods follow the single frame paradigm, and the P4 feature \cite{petr} is leveraged by default. Note that test time augmentation is not used in our method.

Table.~\ref{tab:val_tab} shows the comparison between state-of-the-art methods in nuScenes val set. Both the results with ResNet-50 and ResNet-101 are evaluated. Specifically, with ResNet-50, our 3DPPE achieves 0.370 mAP and 0.433 NDS, outperforming PETR by 3.1\%  and 3.0\%.
When using a stronger ResNet-101 backbone, the performance of 3DPPE boosts to 0.391 mAP and 0.458 NDS, performing better than other competitors. This comparison shows the superiority of our point PE against the camera-ray PE.

We also present the results evaluated by the test server in Table. \ref{test_tab}. In this experiment, we follow the common practice to exploit the DD3D pre-trained VoVNet-99 models. Both the train and val sets are involved in the training phase. Remarkably, 3DPPE achieves 46.0\% mAP and 51.4\% NDS, exceeds PETR by absolute 1.9\% mAP and 1.0\% NDS.

As shown by the results in both tables, the advantage of our 3DPPE in mAP is most pronounced. As the mAP calculation of nuScenes is closely related to the distance to the ground-truth object center, 
this finding further demonstrates that 3DPPE is capable of more precise positioning.

\subsection{Ablation Study} \label{sec:exp_ablation}

We conduct ablative experiments to study the effect of each component in our method. All of the experiments are performed without CBGS strategy. We use C5 feature out of ResNet-50 as the image feature. The resolution of input images is set to $704 \times 256$ by default.

\noindent
\textbf{Effectiveness of the Depth Quality.} 
In our hybrid-depth module,
smooth L1 loss $L_{\mathtt{smooth}-\mathtt{L1}}$ and distribution focal loss $L_{\mathtt{dfl}}$ are adopted as training objectives of the regression depth and the classification depth, respectively. As shown in Table.\ref{tab:tab3}, without any depth supervision, the baseline model achieves 0.343 NDS and 0.266 mAP.
$L_{\mathtt{smooth}-\mathtt{L1}}$ improve the model by 2.2\% NDS and 2.9\% mAP. By involving $L_{\mathtt{dfl}}$,  the performance further boosts to 0.368 NDS and 0.299 mAP. The improved depth quality will provide a more accurate localization for the 3D point positional encoding, verifying the potential of the proposed encoding method.

\begin{table}[htb] 
    \setlength{\tabcolsep}{4.1mm}
    \caption{Effect of Losses in our hybrid-depth head. The results reported in this table are evaluated on nuScenes val set. By default, the backbone network is ResNet-50, and the resolution of input images is $704\times256$.}
    \centering
    \footnotesize
    \vspace{0.25cm}
    \begin{tabular}{c|c|ccc}
    \toprule[1.5pt]
    $L_{\mathtt{smooth}-\mathtt{L1}}$ & $L_{\mathtt{dfl}}$ & NDS$\uparrow$ & mAP$\uparrow$ & mATE$\downarrow$
    \\
    \noalign{\smallskip}
    \hline
    \noalign{\smallskip}
               &          &0.343&0.266&0.832 \\
    \checkmark &          &0.365&0.295&0.818 \\
    \checkmark &\checkmark&0.368&0.299&0.807 \\
    \bottomrule[1.5pt]
    \end{tabular}
    \label{tab:tab3}
\end{table}

\noindent
\textbf{Comparison of 3D Postion-aware Feature}. PETR series and our method all tend to transform 2D image feature to 3D position-aware feature, as such 3D representation can be directly integrated into query-based method for 3D object detection. We aim to demonstrate that the depth-guided 3D point PE is most effective way to construct the 3D position-aware feature, and 5 paradigms of positional encoding listed in the second row of Table.~\ref{tab:tab4} are used for comparison. The depth-guided 3D point PE achieves superior performance compared to the PE in PETR series, it outperforming camera-ray in PETR by 3.1\% NDS and 2.5\% mAP, and surpasses the feature-guided scheme in PETRv2 with 1.6\% NDS and 1.6\% mAP.

\begin{table}[!htb] 
    \caption{Comparison of different 3D position-aware feature. 
    Camera-ray and feature-guided (extended version of camera-ray) are proposed in PETR and PETRv2 respectively.
    The last three rows in point-aware scheme are proposed ourselves:
    topk-aware method selects 5 depth bins with highest probability; 
    depth feature-guided category involves the depth feature in the positional encoding; 
    Depth-guided 3D point approach transforms pixels on image plane to 3D space with predicted depth for precise location.
    }
    \setlength{\tabcolsep}{2.8mm}
    \centering
    \footnotesize
    \vspace{0.25cm}
    \begin{tabular}{lc|p{0.6cm}p{0.6cm}p{0.8cm}}
    \toprule[1.5pt]
    \multicolumn{2}{l|}{3D Position-aware Feature} & NDS$\uparrow$ & mAP$\uparrow$ & mATE$\downarrow$ \\
    \noalign{\smallskip}
    \hline
    \noalign{\smallskip}
    \multirow{2}{*}{{\begin{tabular}[c]{@{}c@{}}Ray-\\aware\end{tabular}}}
    & Camera-Ray \cite{petr} &0.337&0.274&0.852\\
    & Feature-guided \cite{petrv2} &0.352&0.283&0.843\\
    \hline
    \multirow{3}{*}{{\begin{tabular}[c]{@{}c@{}}Point-\\aware\end{tabular}}}
    & Topk-aware            &0.327&0.265&0.869\\ 
    & Depth Feature-guided  &0.359&0.291&0.826\\
    & Depth-guided 3D Point &0.368&0.299&0.807\\
    \bottomrule[1.5pt]
    \end{tabular}
    \label{tab:tab4}
\end{table}

\noindent
\textbf{Effect of Shared 3D Point PE Encoder}. 
This study seeks to provide empirical evidence that the incorporation of unified positional encoding, within the sympatric representation, enhances the detection capacity of 3D objects by the query. In order to assess this proposition, we conducted an experiment, details of which are furnished in Table.~\ref{tab:tab5}, wherein we manipulated the 3D point PE encoder, by varying between a shared and a separate configuration. Our findings demonstrate that the shared positional encoding methodology outperformed the separate approach by a margin of 0.6\% NDS and 0.5\% mAP, demonstrating the effectiveness of our proposed method.

\begin{table}[!htb] 
    \setlength{\tabcolsep}{4.1mm}
    \caption{Effect of separated and shared embedding generator. Shared embedding generator encourages consistency between PE representations and object queries.}
    \vspace{0.25cm}
    \centering
    \footnotesize
    \begin{tabular}{l|ccc}
    \toprule[1.5pt]
    Embedding Generator & NDS$\uparrow$ & mAP$\uparrow$ & mATE$\downarrow$\\
    \noalign{\smallskip}
    \hline
    \noalign{\smallskip}
     Separated &0.362&0.294&0.813\\
     Shared    &0.368&0.299&0.807 \\
    \bottomrule[1.5pt]
    \end{tabular}
    \label{tab:tab5}
\end{table}

\subsection{Qualitative Comparison}\label{sec:exp_showcase}

\begin{figure}[t]
  \centering
   \includegraphics[width=1.0\linewidth]{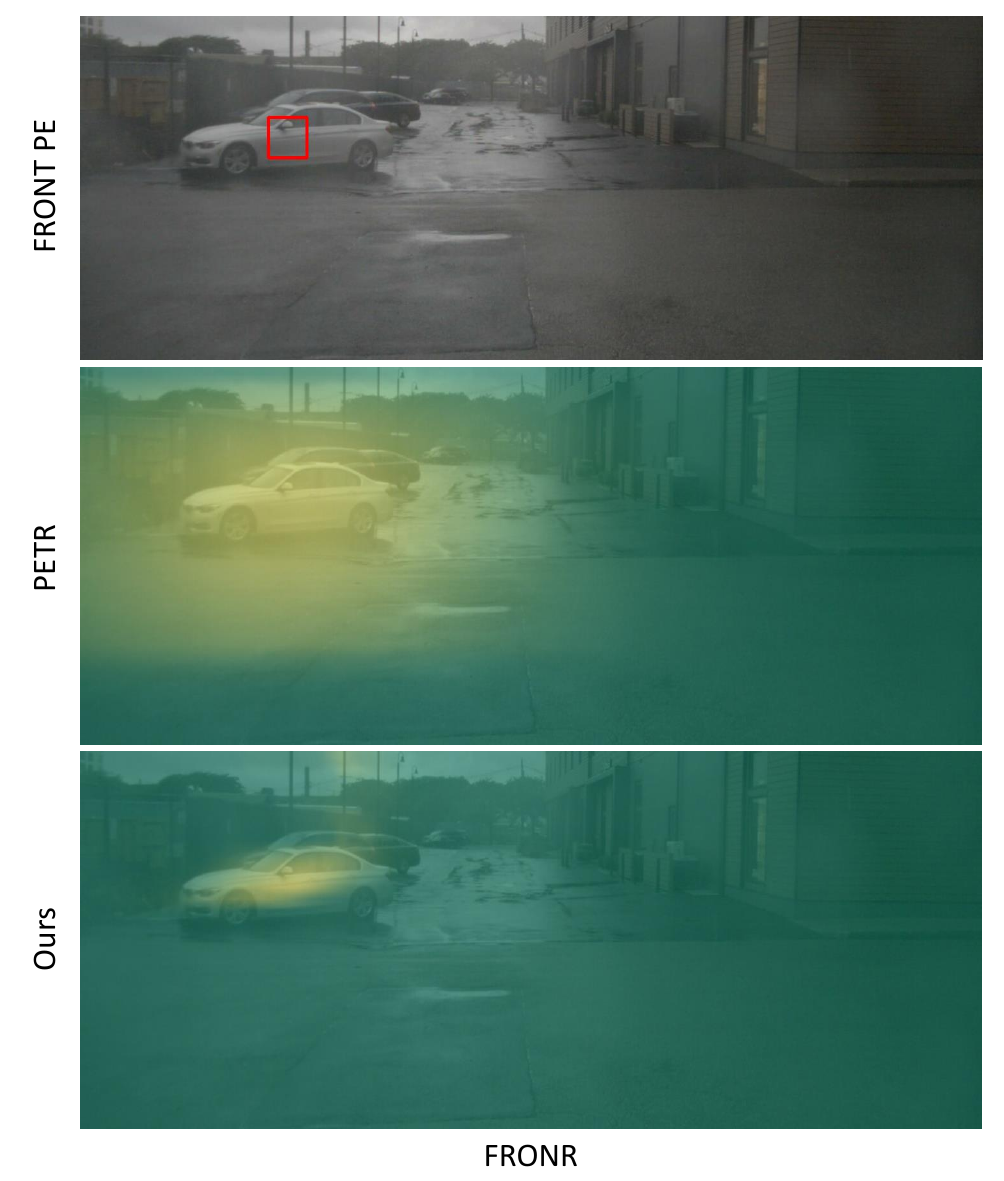}
   \caption{Qualitative comparison of 3D point PE and camera-ray PE in terms of the local similarity (best viewed in color). 
   The red box of the first line indicates a selected pixel. 
   }
   \label{fig:fig5}
  
\end{figure}

We randomly select a pixel from the front view, and compute the similarity of the selected position with all surrounding pixels. 
We find that the similarity of 3D point PE among positions in the local region is higher than that of camera-ray PE, as shown in Figure. \ref{fig:fig5}, yellow region of the former is more cohesive compared to the latter. This indicates that 3D point PE is capable of more precise positioning.

\subsection{Discussion on Further Improvements} \label{sec:exp_discuss}

Our proposed 3DPPE can serve as a simple yet effective baseline, which can be easily extended to achieve better performance. Here, we discuss the potential improvements by (a) leveraging temporal coherence and (b) reusing ground-truth depth for knowledge distillation.

\noindent \textbf{Leveraging Temporal Coherence.} Similar to the extension of PETR (\emph{i.e.}, PETR-v2), our 3DPPE can perform temporal modeling by making use of more frames and taking 3D coordinates calibration. As shown in Table.~\ref{tab:temporal}, 
to validate that our 3DPPE is effective even the backbone is initialized with parameters pretrained on depth estimation tasks, we select VoVNet-99 as the backbone network.
The original temporal coherence modeling in PETR-v2 improves the baseline with 7.6\% NDS and 3.3\% mAP. The gain increases to 8.4\% NDS and 3.9\% mAP when the temporal coherence modeling is transferred to our 3DPPE. Furthermore, our temporal coherence 3DPPE boosts the temporal coherence PETR-v2 with 2.2\% NDS and 2.2\% mAP. 
The observed results indicate the potential of our 3DPPE as an effective extension for temporal coherence.

\begin{table}[htb] 
    \caption{Results of leveraging temporal coherence by involving multiple frames. Here, we exploit VoVNet-99 as the backbone network and set the input resolution as $800\times320$, P4 feature is selected as 2d image feature. By default, 2 frames are used if a model is with temporal information.}
    \centering
    \footnotesize
    \vspace{0.10cm}
    \setlength{\tabcolsep}{16pt}
    
    \begin{tabular}{l|c|cc}
    \toprule[1.2pt]
    Method & Temporal & mAP$\uparrow$ & NDS$\uparrow$
    \\
    \noalign{\smallskip}
    \hline
    \noalign{\smallskip}
    PETR-v2           &        & 0.377 & 0.426 \\
    PETR-v2           &    \checkmark & 0.410 & 0.502 \\
    \midrule[0.8pt]
    3DPPE &          & 0.393 & 0.440 \\
    3DPPE &\checkmark& 0.432 & 0.524 \\
    \bottomrule[1.2pt]
    \end{tabular}
    \label{tab:temporal}
    \vspace{-0.25cm}
\end{table}

\noindent \textbf{Reusing GT Depth for Knowledge Distillation.} 
We show that besides directly exploiting the ground-truth depth as the supervision of our depth estimation network, we can also reuse it for knowledge distillation to achieve further model boosting. The experimental results are shown in Table.~\ref{tab:kd}. Specifically, we first train a 3DPPE model with ground-truth depth, as discussed in Section.~\ref{sec:3.3}. This model is denoted as 3DPPE-oracle. With VoVNet-99 backbone and $800\times320$ input resolution, 3DPPE-oracle achieves 0.4740 NDS and 0.4493 mAP. Inspired by~\cite{chen2022d}, to obtain the distilled model 3DPPE-distill, we add an auxiliary branch sibling to the original transformer decoder at the training stage. The parameters of the auxiliary branch are shared with the original transformer decoder, but the reference points are initialized from the 3DPPE-oracle and will not be finetuned during the training phase. This auxiliary branch of 3DPPE-distill follows the same target assignment as that of 3DPPE-oracle at each iteration and is supervised by the ground-truth boxes together with the predicted result out of 3DPPE-oracle. As shown in this table, 3DPPE-oracle boosts the original 3DPPE to 0.454 NDS and 0.397 mAP, which further validates the extension potential of our method.

\begin{table}[htb] 
    \caption{Results of reusing the ground-truth depth for knowledge distillation. Here, we exploit VoVNet-99 as the backbone network and set the input resolution as $800\times320$, P4 feature is selected as 2d image feature.}
    \centering
    \footnotesize
    \vspace{0.10cm}
    \setlength{\tabcolsep}{24pt}
    
    \begin{tabular}{l|cc}
    \toprule[1.2pt]
    Method   & NDS$\uparrow$ & mAP$\uparrow$
    \\
    \noalign{\smallskip}
    \hline
    \noalign{\smallskip}
    3DPPE-oracle     &        0.474 & 0.449 \\
    \midrule[0.8pt]
    3DPPE     &   0.440 & 0.393 \\
    
    3DPPE-distill  & 0.454 & 0.397 \\
    \bottomrule[1.2pt]
    \end{tabular}
    \label{tab:kd}
    \vspace{-0.15cm}
\end{table}

\section{Conclusion}
In this paper, we analyze the formulation of positional encoding that maps 2D image feature into 3D representation. 
We revisit various positional encoding designs and show that 3D point PE encoded from precise 3D point location is vital to multi-camera 3D object detection. 
Capitalizing on the hybrid-depth module for precise positioning, our proposed 3DPPE achieves state-of-the-art performance among single-frame methods. Moreover, we also demonstrate extension potential of our method on leveraging temporal coherence and reusing ground-truth depth for knowledge distillation.
We hope the proposed depth-guided 3D point PE can serve as a strong baseline for 3D perception.

{\small
\bibliographystyle{ieee_fullname}
\bibliography{egbib}
}

\clearpage
\appendix

{\section*{Appendices}\huge }

\section{Dataset and Metric}
 
\noindent\textbf{Dataset.} we conduct experiments on nuScenes dataset, a comprehensive autonomous driving dataset that encompasses a variety of perception tasks, such as detection, tracking, and LiDAR segmentation. The nuScenes dataset comprises 1,000 distinct driving scenes, divided into three distinct subsets for training (700), validation (150), and testing (150) purposes, respectively. Each of these driving scenes includes 20 seconds of perceptual data that are annotated with a keyframe at a frequency of 2 Hz. The data collection vehicle employed in this study is equipped with one LiDAR, five radars, and six cameras that capture a surround view of the vehicle's environment.

\noindent\textbf{Metrics.}
We follow the official protocol to report the nuScense Score (NDS), mean Average Precision (mAP), along with five true positive metrics including mean Average Translation Error (mATE), mean Average Scale Error (mASE), mean Average Orientation Error (mAOE), mean Average Velocity Error (mAVE) and mean Average Attribute Error (mAAE). 
\section{Experimental Details}
For comprehensive comparison, we have conducted experiments with ResNet-50, ResNet-101 and VoVNet-99 as the backbone networks in our experiments. Following the setting of the PETR series, we use P4 feature by default. Specifically, P4 feature is obtained by upsampling the C5 feature (output of the 5th stage) and fused with the C4 feature (output of the 4th stage).
The P4 feature with 1/16 input resolution or the C5 feature with 1/8 input resolution is used as the 2D feature. 
The monocular depth ranges from 0 to 61m. The region of 3D perception space is set to $[-61.2m, 61.2m]$ for $X$ and $Y$ dimension and $[-10m, 10m]$ for $Z$ dimension. The 3D coordinates in point cloud are normalized to [0,1]. As for the hyper-parameters in each loss component, we set $\lambda_{sl1}$/$\lambda_{DFL}$/$\lambda_{cls}$/$\lambda_{reg}$ to be $0.25$/$0.25$/$2.0$/$1.0$/ respectively, and $\lambda_{cls}$ and $\lambda_{reg}$ is the loss weight for classification and regression follow PETR series. AdamW \cite{adamw} optimizer with a weight decay of 0.01 is used for training model, and the learning rate is initialized as 2.0e-4 and decayed with cosine annealing scheme \cite{cosine}. Unless otherwise stated, all experiments with a batch size of 8 are trained for 24 epochs on 4 Tesla V100 GPUs. Test augmentation methods are not used during the inference.

\section{Analysis of 3D Positional Encoding}

\begin{figure}
\centering
		\includegraphics[width=0.9\linewidth]{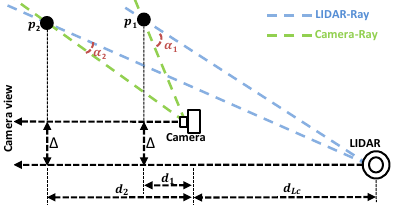}
	\caption{
 A mathematic model for the included angle $\alpha$ between camera-ray and LiDAR-ray in a surround-view system. 
 $d_{Lc}$ is the distance along the camera ray from LIDAR to the camera, 
 $\Delta$ is the distance perpendicular to camera ray from LIDAR to the camera, 
 Apparently, $\alpha_2$ will be smaller than $\alpha_1$ (approaching 0), and meanwhile $d_2$ becomes larger than $d_1$.
	}
	\label{fig:ray_cos}
\end{figure}

\label{sec:analysis_of_position_embbedding}
In this section, we first decoupled the positional encoding of PETR into three factors, \emph{i.e.}, depth values number $N_{\mathtt{D}}$, discretization method $M_{\mathtt{D}}$ and depth range $R_{\mathtt{D}}$. Then, the influence of each factor is explored through ablation studies in Sec.~\ref{sec:3.1}. According to the experimental results, we summarize a feasible physical model to explain the meaning of the positional encoding in PETR.
In Sec.~\ref{sec:3.2}, a new assumption of using LiDAR-ray as positional encoding is proposed. Extensive experiments provide evidence for the new assumption.
All experiments in this section are performed without CBGS and the backbone is set to ResNet-50, C5 feature is selected as 2D image feature, the train image size is set to $704\times256$ defaultly.




\subsection{3D Camera-Ray PE} 
\label{sec:3.1}

PETR series divide the depth range $R_{\mathtt{D}}$ [1m, 61m] into $N^{\mathtt{D}}=64$ depth bins following linear-increasing discretization (LID). Therefore,  one pixel corresponds to 64 separated 3d points lying on the corresponding camera ray. The 3D coordinates of these points are fed together into a 3D positional encoding encoder to generate the PE. In order to clarify what location information is encoded in 3D PE,
we further explore the effectiveness of different depth bin numbers $N_{\mathtt{D}}$, discretization methods $M_{\mathtt{D}}$ and depth range $R_{\mathtt{D}}$ as shown in Table~\ref{tab:tab1}.
\begin{table}[t] 
    \setlength{\tabcolsep}{3.3mm}
    \centering
    \footnotesize
        \caption{
    Quantitative comparison of different depth values number $N_{\mathtt{D}}$, discretization methods $M_{\mathtt{D}}$ and depth range $R_{\mathtt{D}}$. $SID$ and $UD$ denote spacing-increasing discretization and uniform discretization respectively. 
    The invariable performances of top 2-4 rows with diverse $M_{\mathtt{D}}$ indicate that $M_{\mathtt{D}}$ is irrelevant. 
    Thus we fix $M_{\mathtt{D}}$ as simplest $UD$ but change $R_{\mathtt{D}}$ as in row 5 and 6, consistent performances demonstrate that $R_{\mathtt{D}}$ is also incoherence. 
    Finally, we fix $M_{\mathtt{D}}$ and $R_{\mathtt{D}}$ but reduce the $N_{\mathtt{D}}$ to 32 and 2 respectively in last 2 rows,
    the immune performances declare that $N_{\mathtt{D}}$ also does not largely affect the results.  
    }
    \vspace{0.25cm}
    \begin{tabular}{c|c|c|ccc}
    \toprule[1.5pt]
    $N^{\mathtt{D}}$ & $M_{\mathtt{D}}$ & $R_{\mathtt{D}}$ & NDS$\uparrow$ & mAP$\uparrow$ & mATE$\downarrow$ \\
    \noalign{\smallskip}
    \hline
    \noalign{\smallskip}
    64 & $LID$ &    [1,61] &0.338&0.275&0.853 \\ 
    64 & $SID$ &    [1,61] &0.343&0.275&0.856 \\
    64 & $UD$ &     [1,61] &0.337&0.273&0.847 \\
    64 & $UD$  &    [1,31] &0.340&0.274&0.855 \\
    64 & $UD$  &   [31,61] &0.336&0.272&0.849 \\
    32 & $UD$ &     [1,61] &0.342&0.274&0.857 \\
     2 &   -  &     [1,61] &0.345&0.276&0.845 \\
    \bottomrule[1.5pt]
    \end{tabular}

    \label{tab:tab1}
\end{table}
Surprisingly, the results turn out to be almost invariable under different settings, where the fluctuations of $NDS$, $mAP$ and $mATE$ are smaller than 0.8\%, 0.4\% and 1.0\% respectively. 
It gives a intuitive hint that the performance remains virtually unchanged through the separated 3D points sliding on the camera-ray.
If the sampled points can represent the direction of the camera-ray, it already provides equivalent information to the PETR's 3D encoding. Thus, we propose a 3D camera-ray assumption that we could encode the 2D feature by two points on the camera-ray penetrating this pixel.


\subsection{LiDAR-Ray PE Assumption}
\label{sec:3.2}

In this section, we further reduce $N_{\mathtt{D}}$ to 1 with fixed depth $d$ such as 0.2m, 1m, 15m, 30m and 60m respectively.
As listed in Table~\ref{tab:N^D==1}, 
smaller $d$ leads to inferior performance (row 1 and 2 in Table \ref{tab:N^D==1}).
It indicates that the scheme of current PE is no longer camera-ray,
while on-par result (compared to results listed in Table \ref{tab:tab1}) is achieved when $d$ is larger than 15m.
The phenomenon above enlightens us that the PE in PETR with $N^{\mathtt{D}}=1$ represents a LiDAR-ray.
Determination of a ray direction requires two points. As the fixed LiDAR's location provides the start point of ray, thus we can determine LIDAR-ray direction with one point.

As shown in Figure \ref{fig:ray_cos} (b), we calculate the discrepancy ($\mathtt{Dis}$) between camera-ray and LiDAR-ray with the cosine of their included angle:
\begin{align}
    \begin{split}
    \mathtt{Dis} &= 1 - \mathtt{cos}(\alpha) \\
    &= 1 - \mathtt{cos}(\alpha_c - \mathtt{arctan}(\frac{\mathtt{tan}\alpha_c+\frac{\Delta}{d}}{1+\frac{d_{L_c}}{d}})) \\
    &\approx 0.0 \quad \mathtt{when}~~\,d \gg d_{L_c} \mathtt{and}~~ d \gg \Delta,
    \end{split}
\label{eq:included angle cosine}
\end{align}
where $\alpha_c$ is the azimuth angle of camera-ray, $\alpha$ is the included angle between camera-ray and LiDAR-ray, $d_{L_c}$ is the distance between camera and LiDAR along camera view (which ranges from 0.5m to 1.2m in universal sensor configuration), 
$\Delta$ is the distance between camera and LiDAR vertical to camera view
(which ranges from 0.0m to 1.0m in universal sensor configuration). 
When $d \gg d_{L_c}$ and $d \gg \Delta$, the discrepancy between camera-ray and LiDAR-ray is almost eliminated. This further demonstrates that when $d$ is larger, the LiDAR-ray will have a similar direction as the camera-ray, thus, the experiment results will be on par with the original PETR positional encoding even with one point.




\begin{table}[!htb] 
    \setlength{\tabcolsep}{3.7mm}
    \centering
    \footnotesize
    \caption{
    Quantitative comparison of different 
    fixed depth $d$ when depth point number $N_{\mathtt{D}}$ is 1.
    }
    \vspace{0.25cm}
    \small
    \begin{tabular}{c|c|c|ccc}
    \toprule[1.5pt]
    $N^{\mathtt{D}}$ & $M_{\mathtt{D}}$ & $d$ & NDS$\uparrow$ & mAP$\uparrow$ & mATE$\downarrow$ \\
    \noalign{\smallskip}
    \hline
    \noalign{\smallskip}
     1 &   -  &     0.2    &0.304 &0.229 &0.948\\
     1 &   -  &     1      &0.323 &0.251 &0.907\\
     1 &   -  &     15     &0.333 &0.275 &0.842\\
     1 &   -  &     30     &0.340 &0.271 &0.844 \\
     1 &   -  &     60     &0.338 &0.275 &0.849 \\
    \bottomrule[1.5pt]
    \end{tabular}
    \label{tab:N^D==1}
\end{table}



\section{More Ablation Study}
Experiment settings in this section are following setting in the above analysis of 3D positional encoding.

\subsection{More Studies on PE Similarity}
\begin{figure*}[t]
  \centering
   \includegraphics[width=0.75\linewidth]{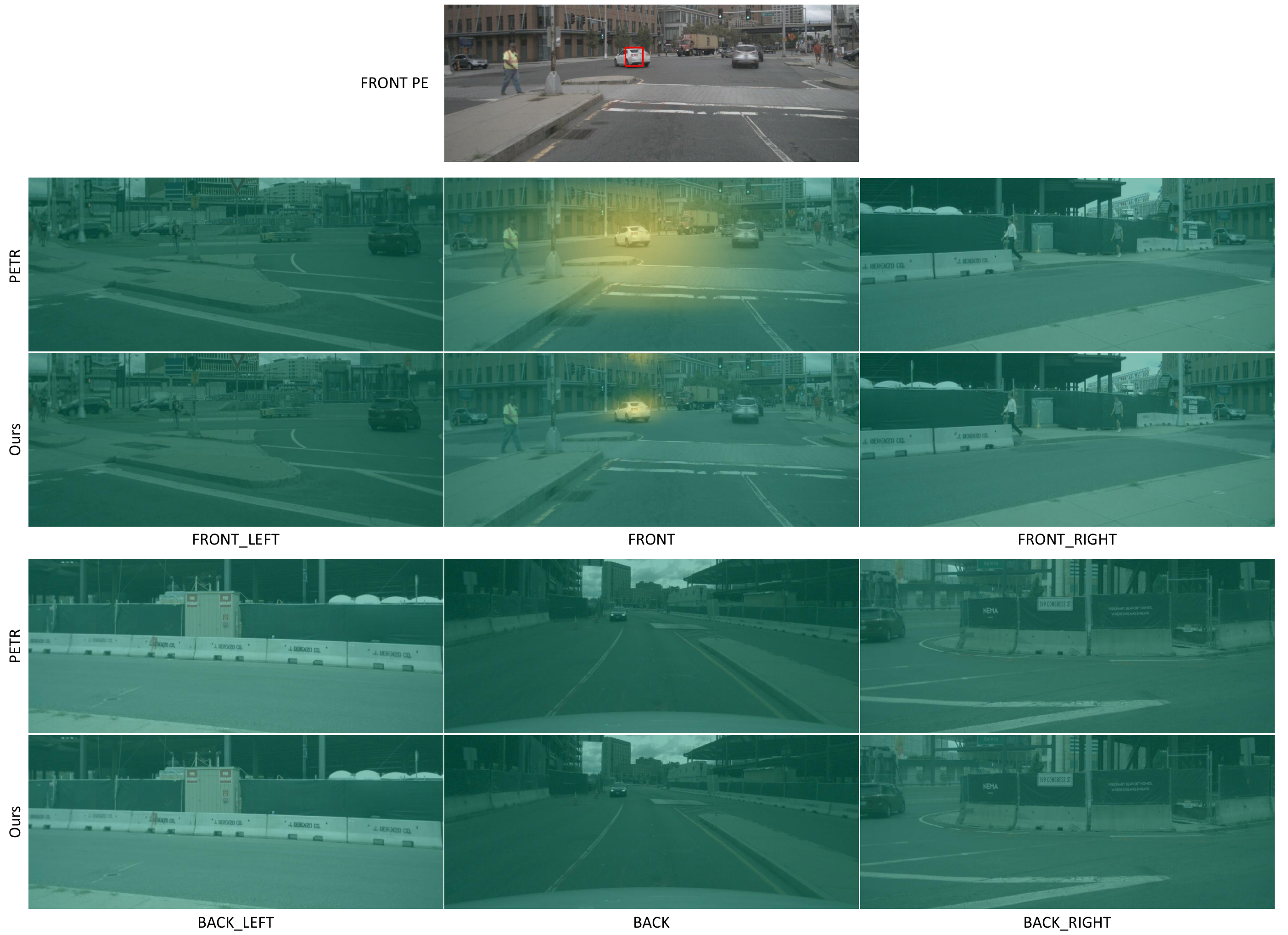}
   \caption{Representative similarity comparison of 3D camera-ray PE in PETR and ours 3D point PE, best viewed in color.}
   \label{fig:fig5-1}
\end{figure*}

\begin{figure*}[t]
  \centering
   \includegraphics[width=0.75\linewidth]{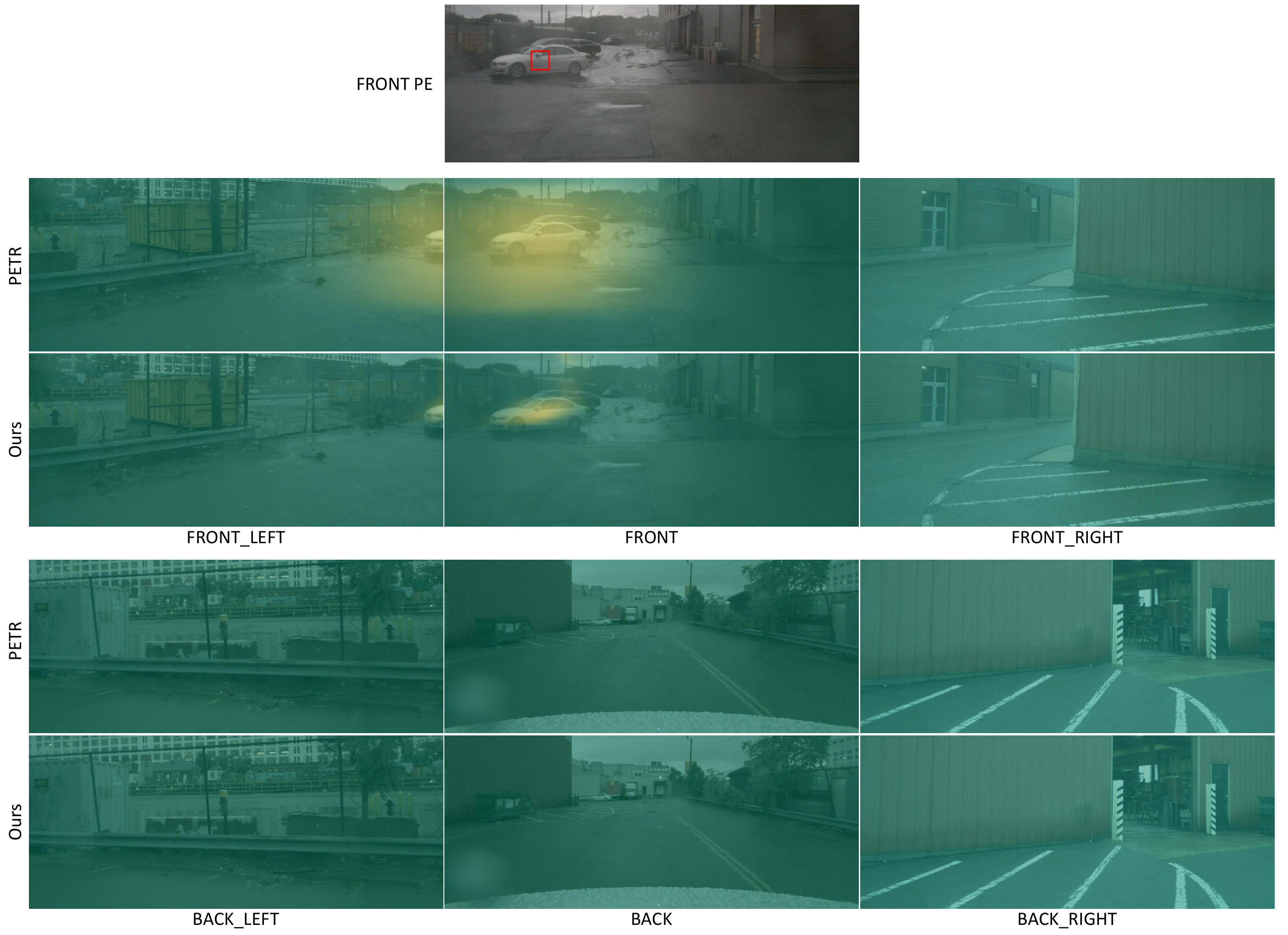}
   \caption{Representative similarity comparison of 3D camera-ray PE in PETR and ours 3D point PE, best viewed in color. The position is selected on the car object appeared in cross-view.}
   \label{fig:fig5-2}
\end{figure*}

\begin{figure*}[t]
  \centering
   \includegraphics[width=0.75\linewidth]{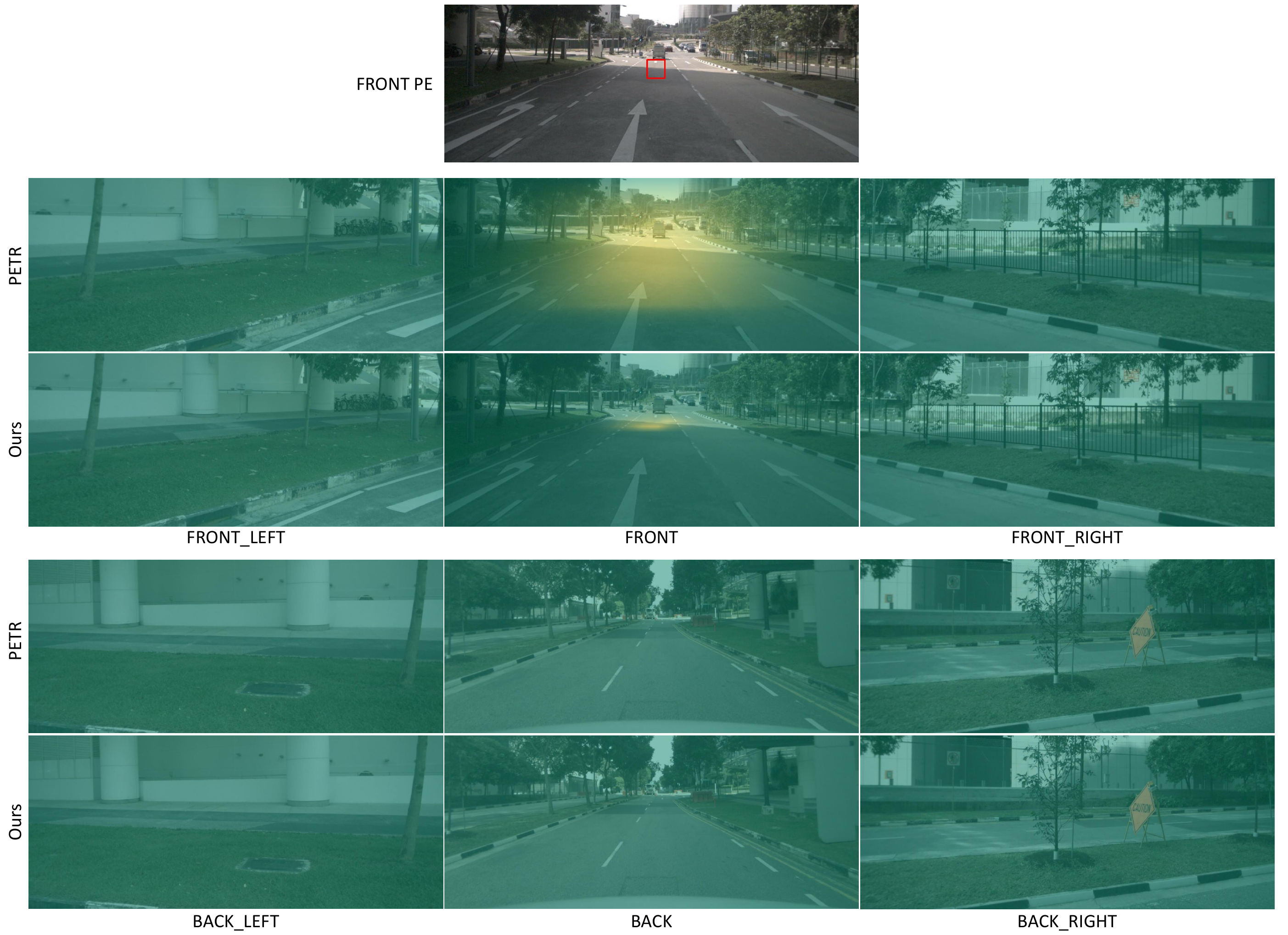}
   \caption{Representative similarity comparison of 3D camera-ray PE in PETR and ours 3D point PE, best viewed in color. The position is selected on the background.}
   \label{fig:fig5-3}
\end{figure*}

To demonstrate the 3D point PE is capable of more precise locating capability, 
we randomly select position on background and object (appeared on the cross-view) respectively from the front view, the similarity between the position and all pixels of the surround views is computed. Figure \ref{fig:fig5-1} is the complete version of the Figure 6 in main paper.
In terem of point selected on cross-view object, 
as illustrated in Figure \ref{fig:fig5-2}, the 3D point PE can find the related object from other view without redundant focus on the background. In term of the PE selected at the road, the 3D point PE tend to focus on the closer region round the selected position compared to the 3D camera-ray PE, as vividly shown in Figure \ref{fig:fig5-3}.

\begin{figure*}[t]
  \centering
   \includegraphics[width=0.95\linewidth]{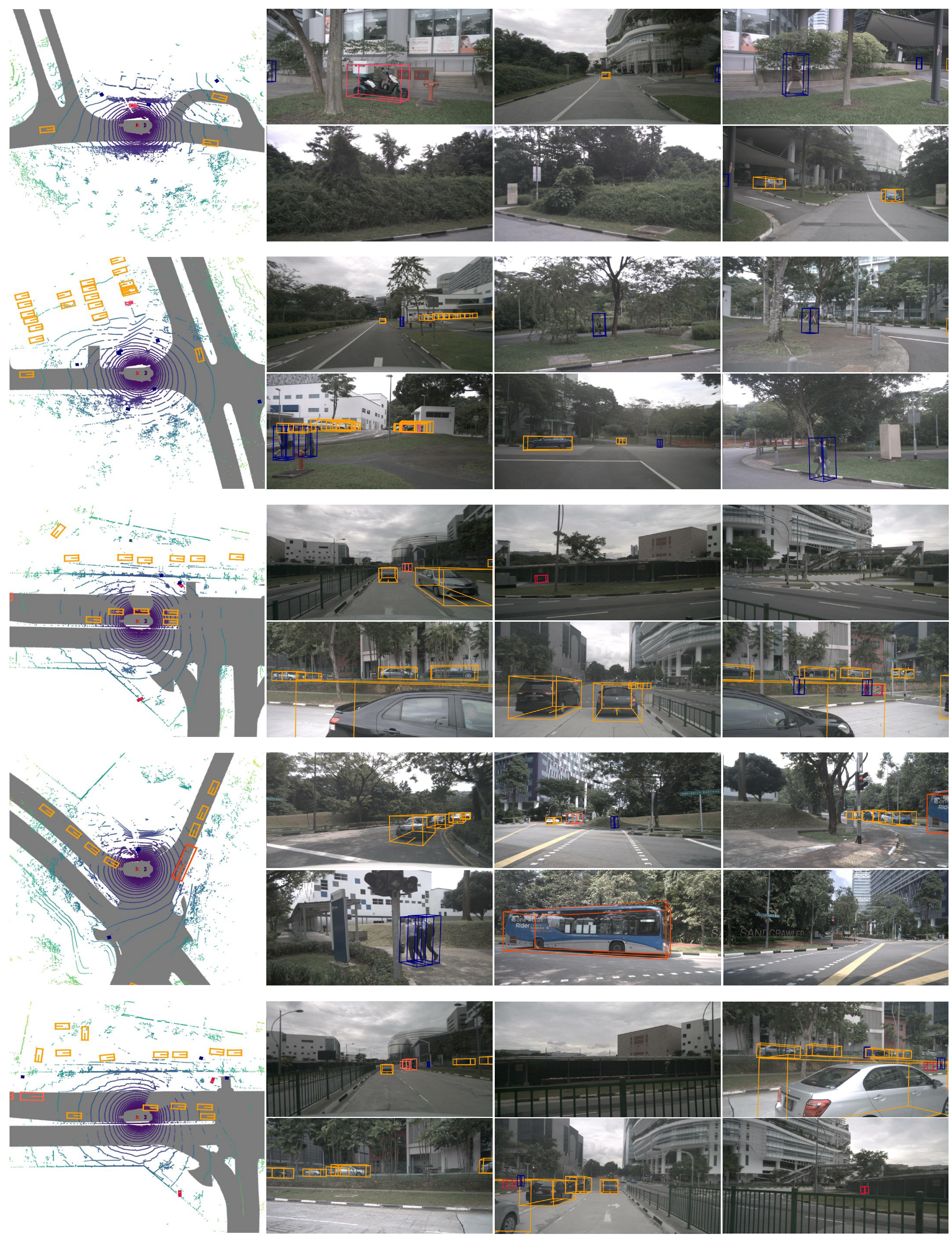}
   \caption{Qualitative results for our method, best viewed in zoom and color.}
   \label{fig:fig6}
\end{figure*}

\end{document}